\documentclass[conference]{IEEEtran}
\IEEEoverridecommandlockouts
\usepackage{cite}
\usepackage{amsmath,amssymb,amsfonts}
\usepackage{algorithmic}
\usepackage{graphicx}
\usepackage{textcomp}
\usepackage{xcolor}
\usepackage{multirow}
\usepackage{bm}
\usepackage{amsmath}
\usepackage[hyphens]{url}
\usepackage{breakurl}

\def\BibTeX{{\rm B\kern-.05em{\sc i\kern-.025em b}\kern-.08em
    T\kern-.1667em\lower.7ex\hbox{E}\kern-.125emX}}
    
\usepackage[acronym]{glossaries}

\newcommand{\traj}{\mathcal{T}}

\usepackage[acronym]{glossaries}

\newacronym{ppo}{PPO}{Proximal Policy Optimization}
\newacronym{mse}{MSE}{mean squared error}
\newacronym{mtp}{MTP}{Motion-To-Photon}
\newacronym{iot}{IoT}{Internet of Things}
\newacronym{drl}{DRL}{Deep Reinforcement Learning}
\newacronym{e2e}{E2E}{end-to-end}
\newacronym{arma}{ARMA}{Autoregressive Integrated Moving Average}
\newacronym{dof}{DoF}{degrees of freedom}
\newacronym{ar}{AR}{Auto-Regressive Model}
\newacronym{tcp}{TCP}{Transmission Control Protocol}
\newacronym{tc}{TC}{Traffic Control}
\newacronym{rmse}{RMSE}{Root Mean Squared Error}
\newacronym{rmp}{RMP}{Riemannian Motion Policy}
\newacronym{pid}{PID}{Proportional Integral Derivative}
\newacronym{aol}{AoL}{age of loop}
\newacronym{aoi}{AoI}{age of information}
\newacronym{voi}{VoI}{value of information}
\newacronym{ros}{ROS}{Robot Operating System}
\newacronym{vr}{VR}{Virtual Reality}

\begin{document}

\title{Real-Time Interactions Between Human Controllers and Remote Devices in Metaverse\\}

\author{
  \IEEEauthorblockN{Kan Chen\textsuperscript{*1},
                    Zhen Meng\textsuperscript{*2},
                    Xiangmin Xu\textsuperscript{*1},
                    Changyang She\textsuperscript{3},
                    Philip G. Zhao\textsuperscript{4}}
\textit{\textsuperscript{1}School of Engineering, University of Glasgow, UK} \\
\textit{\textsuperscript{2}School of Computer Science, University of Glasgow, UK} \\
\textit{\textsuperscript{3}School of Electrical and Information Engineering, University of Sydney, Australia.} \\
\textit{\textsuperscript{4}Department of Computer Science, University of Manchester, UK.} \\
\textit{k.chen.1@research.gla.ac.uk, z.meng.1@research.gla.ac.uk, x.xu.1@research.gla.ac.uk,} \\
\textit{shechangyang@gmail.com, philip.zhao@manchester.ac.uk}
}

\maketitle
\begin{abstract}
Supporting real-time interactions between human controllers and remote devices remains a challenging goal in the Metaverse due to the stringent requirements on computing workload, communication throughput,  and round-trip latency. In this paper, we establish a novel framework for real-time interactions through the virtual models in the Metaverse. Specifically, we jointly predict the motion of the human controller for 1) proactive rendering in the Metaverse and 2) generating control commands to the real-world remote device in advance. The virtual model is decoupled into two components for rendering and control, respectively. To dynamically adjust the prediction horizons for rendering and control, we develop a two-step human-in-the-loop continuous reinforcement learning approach and use an expert policy to improve the training efficiency. An experimental prototype is built to verify our algorithm with different communication latencies. Compared with the baseline policy without prediction, our proposed method can reduce 1) the Motion-To-Photon (MTP) latency between human motion and rendering feedback and 2) the root mean squared error (RMSE) between human motion and real-world remote devices significantly. 

\end{abstract}


\section{Introduction}
\footnotetext {*These authors contributed equally to this work and should be considered co-first authors}
The Metaverse is envisioned as the next generation of the internet, which aims to revolutionize the connections among humans, real-world devices, and virtual objects~\cite{lee2021all}. The virtual objects in the Metaverse can be considered as the digital mirror of the real world which is called ``digital twins"~\cite{glaessgen2012digital}. Digital twins can help to simulate real situations and their outcomes, ultimately allowing them to make better decisions. For example, remote orthopedic surgery enables patients in underserved areas to access specialized surgical expertise without the need for extensive travel. It can combine advanced technologies to replicate the surgeon's skills, enhance visualization, and ensure real-time communication, ultimately improving patient outcomes and expanding the reach of specialized medical care~\cite{laaki2019prototyping}.

\begin{figure}
\centering
\includegraphics[scale=0.31]{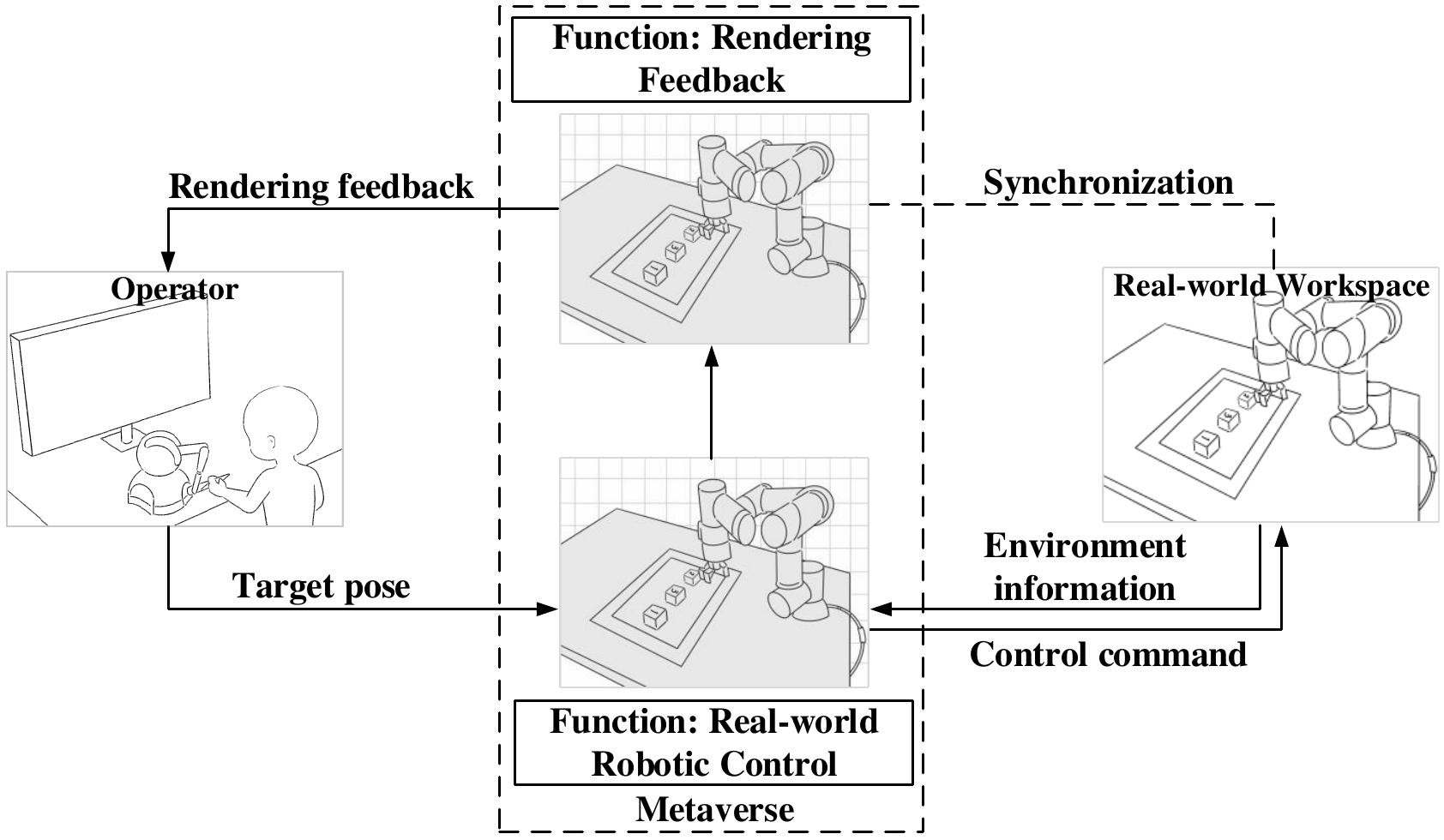}
\caption{Proposed real-time interactions framework for humans, a real robotic arm, and its coupled virtual robotic arm in the Metaverse, where sensing, communication, predication, control, and rendering are considered.}
\label{fig: system model}
\end{figure}

Although the Metaverse realizes more advanced interaction, there are still many challenging problems remaining to achieve the full vision of the Metaverse. First, the increasingly ubiquitous application of AI and growing sophistication of rendering techniques in the Metaverse imposes a substantial burden on computational resources. 
At the same time, various heterogeneous data from different sensors need to be transmitted, thus challenging the data rate, latency, and reliability of the communication system to achieve seamless interconnectivity and immersive user experience~\cite{10422886}. The fine-grained KPIs in VR video streaming have been released in~\cite{3GPP}, where the typical KPIs include 1 Gbps (smooth play) or 2.35 Gbps (interactive) data rate~\cite{mangiante2017vr} to support 8K resolution, 120~FPS, and 360~degree visual field VR video streaming. Besides, the seamless transmission of haptic sensor feedback is also crucial for achieving an immersive experience in the Metaverse. Latency in data transmission is particularly critical for haptic feedback since humans are more sensitive to touch delays than audio or visual ones~\cite{7470948}. In light of this, unlike traditional communication systems designed for high data rates in audio and visual media, the Tactile Internet (TI) prioritizes replicating the sense of touch and kinesthetics~\cite{tactile_she}. To create a seamless experience in teleoperation, the Tactile Internet necessitates ultra-low round-trip latency, ideally below 1~ms~\cite{7470948}. This ensures near-instantaneous responses to user actions, mimicking human reaction times and providing realistic sensing feedback in the virtual world. However, existing 5G network standards struggle to meet the stringent requirements of tactile internet~\cite{tactile_she}. 

Furthermore, the human-in-the-loop control structure in the Metaverse makes eliminating the effects of time delay even more complicated. Delayed control commands and feedback in real-time interactions can significantly degrade the task performance in the Metaverse~\cite{10466554}. Although the effect of latency on human behavior has been initially investigated in teleoperation~\cite{8283715}, the complex interactions and feedback in the Metaverse, as well as more precise quantification, to predict human behavior remains difficult. This further makes it challenging to deploy AI for training and testing. Therefore, how to design a system to improve real-time interaction in the Metaverse is still an open issue.

Significant contributions have been made to reduce the latency in communication systems. Prediction plays an important role in reducing the user-experienced delay in different \gls{vr} systems~\cite{9014097}~\cite{9268977}~\cite{9685224}~\cite{9411714}. In~\cite{9268977}, 
to reduce the latency, the author collaboratively optimized the duration of the observation window for predicting tiles and the duration for computing and transmitting these predicted tiles, aiming to achieve a balance in performance across prediction, communication, and computing tasks to maximize QoE given any arbitrary predictor and configured resources. In~\cite{9411714}, the authors considered the limitations of the computing capacity of end-user devices, where an MEC-enabled wireless VR network is proposed to predict the field of view (FoV) of each VR user by using the Recurrent Neural Network (RNN). However, these efforts only considered the real-time experience of the user on the rendering side and did not take into account the completion of the task in the case of real-time interaction and control. A recent metric named \gls{aoi} aims to quantify the freshness of information and is gradually used to reduce the latency in real-time control scenarios~\cite{NET-060}. In~\cite{9163049}, the authors considered the computation of \gls{aoi} in the case of real-time control with simultaneous sensing, controlling, and actuating. However, these analyses were model-based and lack considerations regarding real-time human interaction systems. It is not feasible for complex systems involving prediction, computation, rendering, and control, especially in the case of human-in-the-loop. Our latest work~\cite{10370739}  predicted the behavior of a operator during real-time control and to reduce the control error of a robotic arm from a task-oriented perspective thereby eliminating the effect of latency. However, the analysis was based on offline data training and testing and did not directly address the latency problem of real-time systems with a human in the loop.

\section{Conribution}

In this paper, we aim to solve the following problems: 1) How to eliminate the effects of time delay among different subsystem components to achieve real-time interactions? 2) How to do the cross-system design, integrating the domain knowledge of the user behavior to solve the human-in-the-loop real-time interaction issue. 3) How to design the prototype to verify the effectiveness of the proposed DRL algorithm. The main contributions of this paper are summarized as follows:

\begin{itemize}
\item We established a novel framework for real-time interactions through the virtual robotic arm in the Metaverse with its coupled real-world remote device, where sensing, communication, prediction, control, and rendering are jointly considered. Specifically, we jointly predict the motion of the human controller for 1) proactive rendering in the Metaverse and 2) generating control commands to the real-world robotic arm in advance. The virtual model is decoupled into two functions for real-world robotic control and rendering feedback, respectively.

\item We developed a two-step human-in-the-loop continuous \gls{drl} approach based on the advanced \gls{ppo} algorithm to adjust the prediction horizons for rendering and control dynamically. We take the expert knowledge, e.g., \gls{aol} as the state for improving the training efficiency.

\item We built a prototype including input device, a real-world robotic arm and its digital model in the Metaverse. Extensive experiments are carried out in the prototype and our proposed method can reduce 1) the \gls{mtp} latency between human motion and rendering feedback and 2) the \gls{rmse} between human motion and real-world remote device significantly.

\end{itemize}

\section{System Model}
As shown in Fig~\ref{fig: system model}, the framework consists of three parts, an operator operating the input device, the Metaverse with a virtual robotic arm, and a real-world workspace with a real robotic arm. The operator and the two workspaces are at a distance, communicating through a wireless channel. Due to communication delay between the operator and the two workspaces, and the delays caused by prediction, computation, and rendering processes, discrepancies exist in the perceived trajectory of the virtual robotic arm, the real-world remote robotic arm, and the operator's control commands. When the operator operates the controller, the commands are sampled and predicted to 1) compensate for the Motion-To-Photon (MTP) latency between human motion and rendering feedback and 2) reduce the \gls{rmse} between human motion and the real-world remote device. 

\begin{figure}
\centering
\includegraphics[scale=0.41]{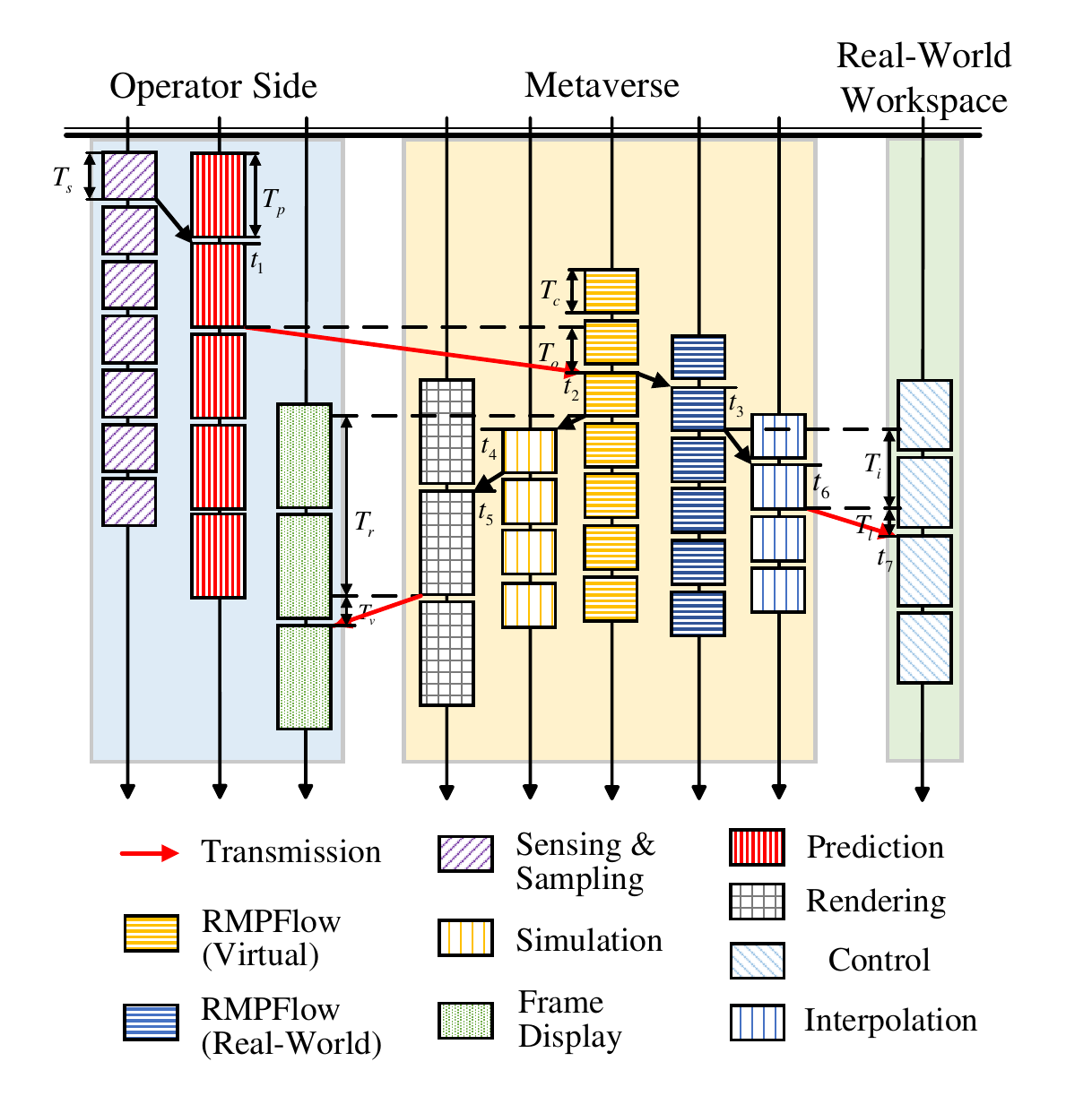}
\caption{The workflow of the proposed framework, where the modeling accuracy and the latency need to be satisfied.}
\label{fig: time flow}
\end{figure}

\subsection{Information Flow}
As shown in Fig~\ref{fig: time flow}, the proposed framework comprises two information flows served for two functions: real-world robotic control and the rendering feedback. The total time duration of real-world robotic control, denoted as $T_{cc}$, encompasses all delays from sensing the movement of the operator through the data transmission, Metaverse data processing, and real-world robotic control. It indicates the time duration from the initiation of a control command by the operator to its execution by the robotics. During the process, the operator operates the input device and generates a series of poses $\{{\bf{p}}_\text{1}, {\bf{p}}_2,...,{\bf{p}}_i\}$ which are sampled by the built-in sensor. This introduces the sensing delay $T_s$. At the same time, we maintain a queue with a fixed length to store these samples. Next, an agent decides the length of the prediction horizon based on the freshest samples $\mathcal{P}(t)$ in the queue, followed by a predictor generates the predicted pose for the rendering $\hat{{\bf{p}}}_r(t)$ and control $\hat{\bf{p}}_c(t)$. Limited by the decision frequency, this process causes a queue delay $T_q$ and a prediction delay $T_p$. Then, $\hat{{\bf{p}}}_r(t)$ and $\hat{{\bf{p}}}_c(t)$ are transmitted to the Metaverse via networks and the queue is emptied. This process introduces the communication delay $T_{o}$. After that, the predicted poses $\hat{{\bf{p}}}_r(t)$ and $\hat{{\bf{p}}}_c(t)$ will be processed in the Metaverse taking into account the simulation or feedback of the environment to generate a joint acceleration control command for the robot and then be interpolated for trajectory smoothing. This will incur a computational delay $T_{c}$ and interpolation delay $T_{i}$. Finally, the command is transmitted to the real-world robotic arm via a communication channel for actuation, incurring a communication delay $T_{l}$. Therefore, the total time duration of the function of real-world robotic control $T_{cc}$ is expressed as 

\begin{align}\label{eq: Tcc}
   T_{cc} =  T_{p} + T_{o} + T_{c} + T_{i} + T_{l}.
\end{align}
On the other hand, the total time duration of the rendering feedback, denoted by $T_{cr}$, encompasses all delays from the sensing of movement of the operator, data transmission, Metaverse data processing, rendering, and feedback display. It indicates the duration from the initiation of a movement by the operator to his perception, which also refers to the \gls{mtp}~\cite{zhao2017estimating}. Specifically, time delays before the step of the Metaverse process are identical to those of real-world robotic control. In addition to the delay of generating control commands for the robotic arm $T_c$, the virtual environment is also simulated and rendered in the Metaverse, which incurred a rendering delay $T_{r}$. Next, the rendered image is streamed back to the user side, which incurs a communication delay $T_{v}$. Therefore, the total time duration of the function of rendering feedback $T_{cr}$ is expressed as,

\begin{align}\label{eq: Tcr}
    T_{cr} =  T_{p} + T_{o} + T_{c} + T_{r} + T_{v}.
\end{align}
In the following subsection, we discuss these sub-components in detail.
\subsection{Operator Side}

On the operator side, we utilize the 3D Touch haptic device~\cite{haptic} as the input device. In the $t$-th time slot, when the operator operates the handle of the 3D Touch, the sampled pose $\mathbf{p}(t)$ is denoted by 
\begin{align}
\mathbf{p}(t) = [l_\mathrm{x}(t), l_\mathrm{y}(t), l_\mathrm{z}(t), q_\mathrm{x}(t), q_\mathrm{y}(t), q_\mathrm{z}(t), q_\mathrm{w}(t)],
\end{align}
where $[l_\mathrm{x}(t), l_\mathrm{y}(t), l_\mathrm{z}(t)]$ is the position of the handle in the Cartesian coordinate system and $[q_\mathrm{x}(t), q_\mathrm{y}(t), q_\mathrm{z}(t), q_\mathrm{w}(t)]$ represents the orientation vector in quaternion~\cite{kuipers1999quaternions}. To compensate for the time delay discussed in Section III.A, we propose to use ~\gls{arma} as the predictor for its low computational complexity, robustness, and no need for pre-training~\cite{}. It is worth noting that our designed framework can be easily replaceable with other prediction algorithms. Specifically, in time slot $t_1$, given historical poses $\mathcal{P}(t_1) = [{\bf{p}}(t_1-W_p), {\bf{p}}(t_1-W_p+1), ...,{\bf{p}}(t_1)]$, ~\gls{arma} predicts poses for real-time robotic control $\hat{\bf{p}}_{\it{c}}(t_1+H_c(t_1))$ and rendering feedback $\hat{\bf{p}}_{\it{r}}(t_1+H_r(t_1))$ respectively. This is expressed by,
\begin{align}\label{eq: prediction}
[\hat{\bf{p}}_{\it{c}}(t_1+H_c(t_1)),&\hat{\bf{p}}_{\it{r}}(t_1+H_r(t_1))] = \notag \\
& \;\;\;\; \mathcal{F}_p(\mathcal{P}(t_1), \theta_{p},H_c(t_1), H_r(t_1)),
\end{align}
where $\theta_{p}$ denotes the parameters of the prediction. $H_c(t_1)$ and $H_r(t_1)$ represent the lengths of prediction horizons for both functions. These lengths are dynamically determined by an agent following the policy $\pi_\theta$ trained by the DRL algorithm. This is expressed by,
\begin{align}
[H_c(t_1), H_r(t_1)] = \pi_\theta([{\bf{p}}(t_1), \Delta_L(t_1)]),
\end{align}
where $\Delta_L(t_1)$ is the \gls{aol}, which is defined as 
\begin{align}
    \Delta_L(t) = t - U(\bf{p}'(\it{t})),
\end{align}
where $\bf{p}'$$(t)$ is the last sampled pose that controls the robotic arm and $U(\cdot)$ is the generation time.

\subsection{The Metaverse}
We establish the Metaverse based on the Nvidia Omniverse Isaac Sim~\cite{sim}. We decouple the two functions; i.e., real-time robotic control and rendering feedback in the Metaverse. After receiving the predicted poses $[\hat{{\bf{p}}}_r(t_1), \hat{{\bf{p}}}_c(t_1)]$ from the operator side, in time slot $t_2$, the control command of virtual robotic arm for two functions are computed separately, which is expressed by,
\begin{align}\label{eq: control_sim}
\widetilde{\traj}_c(t_3) = [\widetilde{\tau}_1(t_3),..., \widetilde{\tau}_I(t_3)] = \mathcal{F}{c}(\hat{{\bf{p}}}_c(t_1),\theta{c}), \\
\widetilde{\traj}_r(t_2) = [\widetilde{\tau}_1(t_2),..., \widetilde{\tau}_I(t_2)] = \mathcal{F}{r}(\hat{{\bf{p}}}_r(t_1),\theta{r}),
\end{align}
where the parameters of both controlling functions are denoted by $\theta_r$ and $\theta_c$, respectively. The target joint angle of the $i$-th robotic arm joint is denoted by $\widetilde{\tau}_i(t)$. Here, we propose to use the RMPFlow control algorithm to compute the joint acceleration for the virtual robotic arm in the Metaverse, which is a highly effective robotic control algorithm that combines local strategies to generate a cohesive global strategy based on the Riemannian Motion Policies (\gls{rmp})~\cite{ratliff2018riemannian, sim_rmp}. Specifically, for the $i$-th joint of a robot, the target acceleration $\ddot{{\bf{q}}}_i(t)$ in the $t$-th time slot for virtual robotic arm actuation is calculated by,
\begin{align}\label{eq: rmpflow}
\ddot{\bf{q}}_i(t) = {\pi_r(\bf{q}(t),\dot{\bf{q}}(t))} = {k_p} {{\mathcal{R}}({\bf{q}}'_i(t) - {\bf{q}}_i(t))} - {k_d} {\dot{\bf{q}}_i(t)},
\end{align}
where ${\bf{q}}'_i(t)$ is the target position vector, $k_p$ represents the position gain determining the strength of configuration pull towards the target, and $k_d$ represents the damping gain determining the amount of resistance. The robust capping function ${\mathcal{R}}({\bf{u}}(t))$ is defined as,
\begin{align}\label{eq: robust capping}
{\mathcal{R}}({\bf{u}}(t)) =  \begin{cases}
{\bf{u}}(t), \ \ \ \ \ \ \ \ \ ||{\bf{u}}(t)|| < \theta \\
\theta{\bf{u}}(t) / ||{\bf{u}}(t)||, \ \ \text{otherwise}
\end{cases}
\end{align}
where the position vector is defined as ${\bf{u}}(t)$, $\theta$ represents the distance at which the position correction vector is capped, and $||\cdot||$ is the Euclidean norm of a position vector. 

Then, the computing results of RMPFlow for real-world robotic control are further interpolated by a trajectory smoothing algorithm~\cite{sim_rmp} for narrowing the gap between sim and real, which is expressed by
\begin{align}\label{eq: control_interpolation}
{\bf{\hat{q}}}_{i}(t) = \begin{cases}
\alpha \cdot {\bf{\hat{q}}}_{i}(t') + (1 - \alpha) \cdot {\bf{q}}_{i}(t), t_e < t_d \\
{\bf{\hat{q}}}_{i}(t'), \ \ \text{otherwise}
\end{cases}
\end{align}
where ${\bf{q}}_{i}(t')$ is the last command and $\alpha$ represents the smooth parameter, which goes linearly from zero to one with quadratic increase. The initialization of $\alpha$ is $\alpha = t_e / t_d $ and the update of $\alpha$ is denoted by $\alpha = \alpha'\cdot\alpha'$.

For the function of rendering feedback, firstly, the virtual environment in the metaverse is simulated discretely based on the minimum simulation step $T_{si}$. In the $j$-th simulation step, the Metaverse simulates the changes in the virtual environment based on the physical law over a certain time, which is denoted by the simulation step length $\Delta_t$. During the simulation process, the virtual robotic arm in the Metaverse can interact with the virtual objects. 

The simulation process in time slot $t_4$ can be represented as
\begin{align}\label{eq: update}
\widetilde{\traj}_s(t_4) = \mathcal{F}_{u}(\widetilde{\traj}_r(t_2),\theta_{u},\theta_{e}),
\end{align}
where $\mathcal{F}_{u}(\cdot, \theta_{u}, \theta_{e})$ represents the rendering process $\theta_{u}$ represents the settings and parameters for physical updates and $\theta_{e}$ represents the interaction with the environment.

Then, we render the image on the Metaverse side and send these frames back to the user side. The rendering process is expressed by
\begin{align}\label{eq: rendering_back}
{\bf{I}}_r(t_5) = \mathcal{F}_{r}(\widetilde{\traj}_r(t_4),\theta_{r}),
\end{align}
where $\mathcal{F}_{r}(\cdot, \theta_{r})$ represents the rendering process and $\theta_{r}$ represents the settings and parameters for rendering. The rendered framed is denoted by ${\bf{I}}_r(t)$. We assume that the processing time for each image is represented by $T_{im}$ time slots, thus the generated video refresh rate is $f_r = 1/(T_s \cdot T_{im})$. These rendering frames are then transmitted back to the user side through the communication system.

The metaverse is also responsible for interpolating the control command $\widetilde{\traj}_c(t_3)$ generated by the RMPFlow controller before sending it to the real-world workspace. This process will smooth the trajectory, which is expressed by
\begin{align}
\widetilde{\traj}_{ci}(t_6) = [\widetilde{\tau}_1(t_6),..., \widetilde{\tau}_I(t_6)] = \mathcal{F}{ci}(\widetilde{\traj}_c(t_3),\theta{ci}),
\end{align}
where $\mathcal{F}_{ci}(\cdot, \theta{ci})$ denotes the interpolation function, whose parameters are denoted by $\theta_{ci}$.

\subsection{Real-world workspace}

In real-world workspace, we adopt UR3e~\cite{ur3e} for real-world robotic control. After receiving the control command $\widetilde{\traj}_{ci}(t_6)$ transmitted from the Metaverse, the movement of UR3e is controlled by the Moveit motion planning framework, where joint position controller is used ~\cite{ros_moveit}.

In the time slot $t_6$, the trajectory of the real robotic arm is expressed by
\begin{align}\label{eq: control_real}
\tilde{\traj}_{rw}(t_7) = \mathcal{F}_{rw}(\widetilde{\traj}_{ci}(t_6),\theta_{rw}),
\end{align}
where $\mathcal{F}_{rw}(\cdot, \theta_{rw})$ is the Moveit motion planning process, and $\theta_{rw}$ denotes the corresponding parameters.

\subsection{Networks}
We propose to use the publish-and-subscribe mechanism in the \gls{ros} in the networks~\cite{ros}, which enables flexible and scalable communication between different sides among operator, Metaverse, and real-world workspace, facilitating modular development and integration of complex Metaverse applications. To better investigate the effect of different communication delays on the experimental results, we connect the control haptic device, server for Metaverse, and the computer controlling the robotic arm via a network cable and communicate via UDP protocol~\cite{ros_udp}. This makes the communication delay negligible thus acting as a baseline. Then, we add different communication delays via the~\gls{tc} method~\cite{tc}.

\section{Problem Formulation}

In this section, we formulate an optimization problem that minimizes the RMSE between the poses of the input device and both robotic arms by optimizing the prediction horizons $H_c(t)$ and $H_r(t)$. Specifically, we propose to use the \gls{ppo} algorithm as the baseline method for its high sampling efficiency and effectiveness~\cite{ppo}. Due to the time-consuming nature of having people in the loop to train DRL, we adopt a two-step \gls{drl} training process, where the DRL is first trained with the recorded pose. Then based on the pre-trained model, the human-in-the-loop model is trained.
\subsection{State}
The state in the $t$-th time slot is set to a combined vector of the latest input pose $\bf{p}$$(t)$ and the \gls{aol} $\Delta_L(t)$, denoted by 
$\bf{s}_{\it{t}} = [{\it{s}^\text{[1]}_t},{\it{s}^\text{[2]}_t}] = [\bf{p}$$(t), \Delta_L(t)], {s}^{[1]}_t \in [-1,1]^{1 \times 7}, {{s}^{[2]}_t} \in \mathbb{N}$.
\subsection{Action}
The action in the $t$-th time slot is the length of prediction of two functions, i.e., real-world robotic control and rendering feedback, which is denoted by ${\bf{a}_t} = [{{{a}}^{[1]}_t},{{{a}}^{[2]}_t}] = [H_r(t), H_c(t)] \in \mathbb{N}^{1 \times 2}$.
\subsection{Reward}
Given $\bf{a}_t$ and $\bf{s}_t$ in the time slot $t$, the instantaneous reward is set to a weighted sum of \glspl{rmse}, which is denoted by
\begin{align}
    r(\textbf{s}_t, \textbf{a}_t) = &\;\;w_1 \times(\bf{RMSE}_p(\bf{o}(\it{t}),\bf{m}(\it{t}))) \\ \notag 
    &+ w_2 \times (\bf{RMSE}_p(\bf{m}(\it{t}),\bf{r}(\it{t}))) \\ \notag
    &+ w_3 \times (\bf{RMSE}_o(\bf{o}(\it{t}),\bf{m}(\it{t}))) \\ \notag
    &+ w_4 \times (\bf{RMSE}_o(\bf{m}(\it{t}),\bf{r}(\it{t}))),
\end{align}
where $\bf{RMSE_p}(\cdot,\cdot)$ is the \gls{rmse} of position and $\bf{RMSE_o}(\cdot,\cdot)$ is the \gls{rmse} of orientation, respectively. The poses of the input device, the virtual model, and the real-world remote robotic arm are denoted by $\bf{o}(\it{t})$, $\bf{m}(\it{t})$ and $\bf{r}(\it{t})$, respectively.

\subsection{Problem Formulation}
The policy ${\pi_{\theta}(\textbf{a}_t \mid {\textbf{s}}_t)}$ is a mapping from the state to different actions, where $\theta$ are the training parameters of the policy network. Following the $\pi_\theta$ policy, the long-term reward can be given by
\begin{equation}
    R^{\pi_\theta} = \mathbb{E}[\sum_{t=0}^{\infty} \gamma^t r(\mathbf{s}_t, \mathbf{a}_t)],
\end{equation}
where $\gamma$ is the reward discounting factor. To find an optimal policy $\pi_\theta^*$ that maximizes the long-term reward $R^{\pi_\theta}$, the problem is formulated as
\begin{align}
   & \;\;\;\;\;\; {\pi_{\theta}^{*}}  =  \mathop {\max }\limits_{\theta} Q^{\pi_{\theta}}({\bf{s}}_t, {\bf{a}}_t)\hfill,    \label{qvalue} \\
    Q^{\pi_{\theta}}({\bf{s}_t}, {\bf{a}_t}) & =  \;{\mathop{\mathbb{E}}}[\sum_{t = 0}^\infty  {{ \gamma ^t}}r({\bf{s}}_t, {\bf{a}}_t) \mid {\bf{s}}_0={\bf{s}},\ {\bf{a}}_0={\bf{a}},\ \pi_{\theta}], \tag{\ref{qvalue}{a}} \\
    & \;\;\; \;\quad \ \; 0<P_r(t) < H_r \tag{\ref{qvalue}{b}} \\
    & \;\;\; \;\quad \ \; 0<P_c(t) < H_c \tag{\ref{qvalue}{c}}
\end{align}
where 
$H_r$ denotes the maximum prediction horizon for the rendering loop and $H_c$ is the maximum prediction horizon for the control loop.

\section{Prototype Design and Results}

\begin{figure}
            \centering
            \includegraphics[scale=0.3]{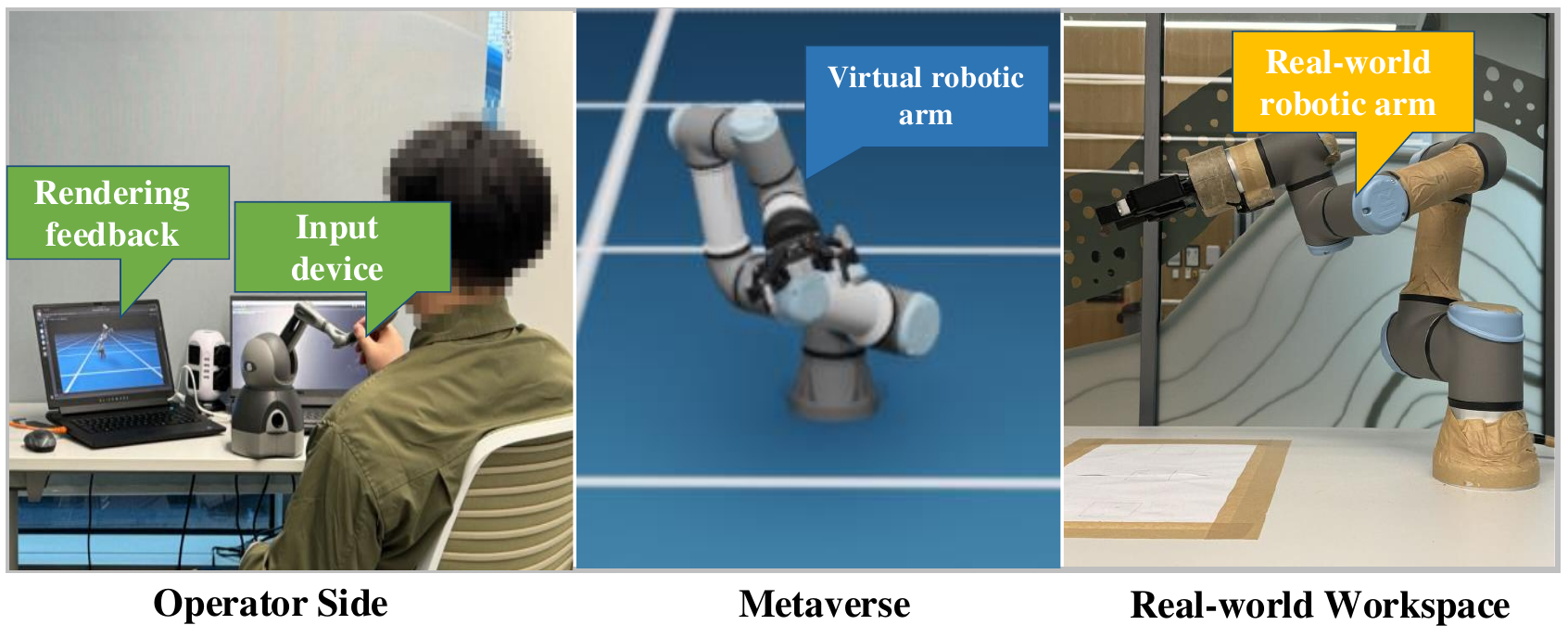}
           \caption{Illustration of our prototype system.}
           
           \label{fig: prototype}
  \end{figure}

In this section, we demonstrate our co-design framework as shown in Fig.\ref{fig: prototype}. Based on the prototype, we first evaluate the performance of prediction model and then evaluate the effectiveness of the proposed framework.

\subsection{Evaluation of the proposed framework}

\begin{table}
\centering
\caption{Evaluation of the Prediction Model}
\label{teble_pre}
\resizebox{0.95\linewidth}{!}{
\begin{tabular}{|c|c|c|c|c|}
\hline
\multicolumn{2}{|c|}{Average \gls{rmse}}  & \gls{arma}    & Informer  & AR \\ 
\hline
\multirow{2}{*}{\begin{tabular}[c]{@{}c@{}} Position ($\text{m}$) 
\end{tabular}}
& Mean                & 0.00100       & 0.00842     &  0.00174 \\ 
\cline{2-5}
&Variance            & $5.264\times10^{-5}$        & 0.00343     &  $1.108\times10^{-4}$ \\ 
\hline
\multirow{2}{*}{\begin{tabular}[c]{@{}c@{}} Orientation 
\end{tabular}}
& Mean                & 0.00101        & 0.01031     &  0.00201 \\ 
\cline{2-5}
&Variance            & $5.650\times10^{-5}$         & 0.00638     &  $0.912\times10^{-5}$ \\ 
\hline
\end{tabular}}
\end{table}

\begin{figure}
\centering
\includegraphics[scale=0.38]{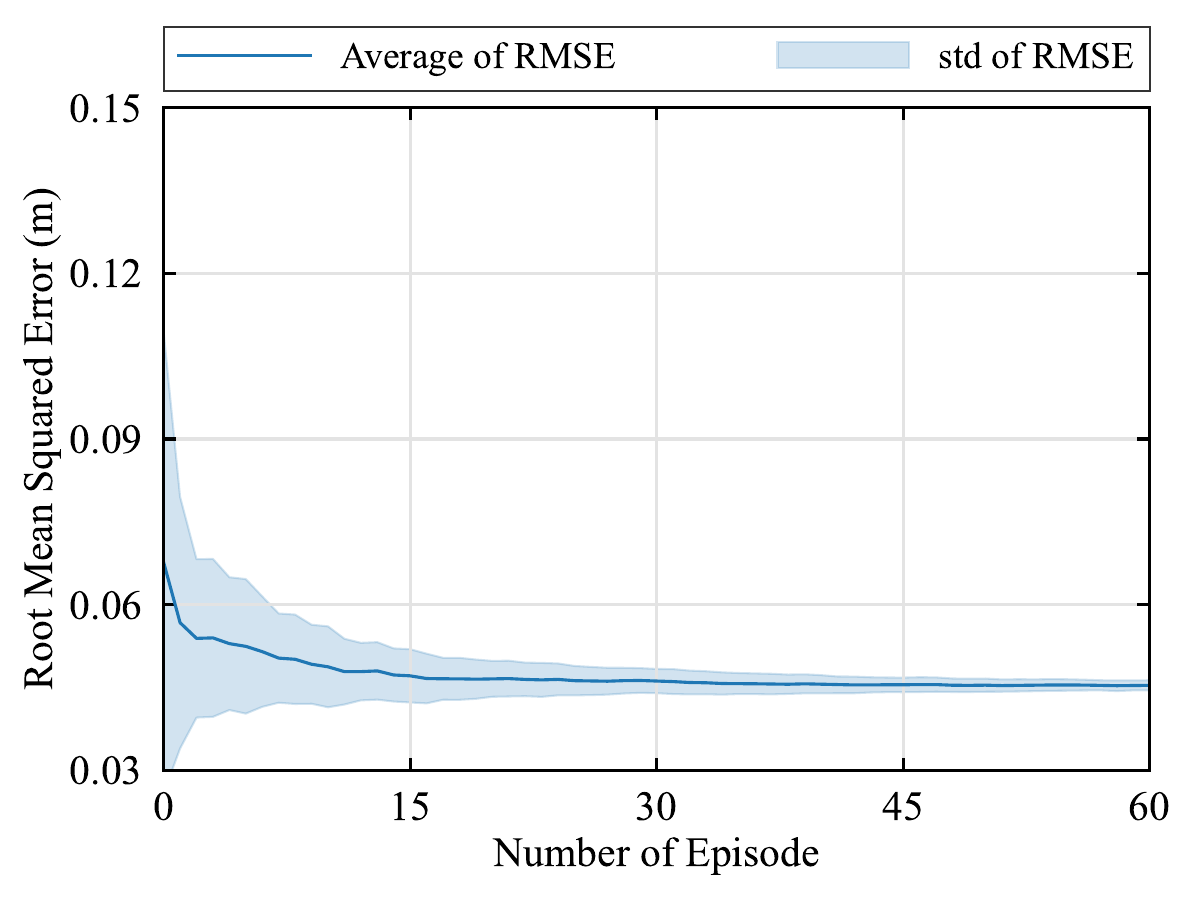}
\caption{Average \gls{rmse} in each training episode.}
\label{fig: reward}
\end{figure}

\begin{figure}
\centering
\includegraphics[scale=0.38]{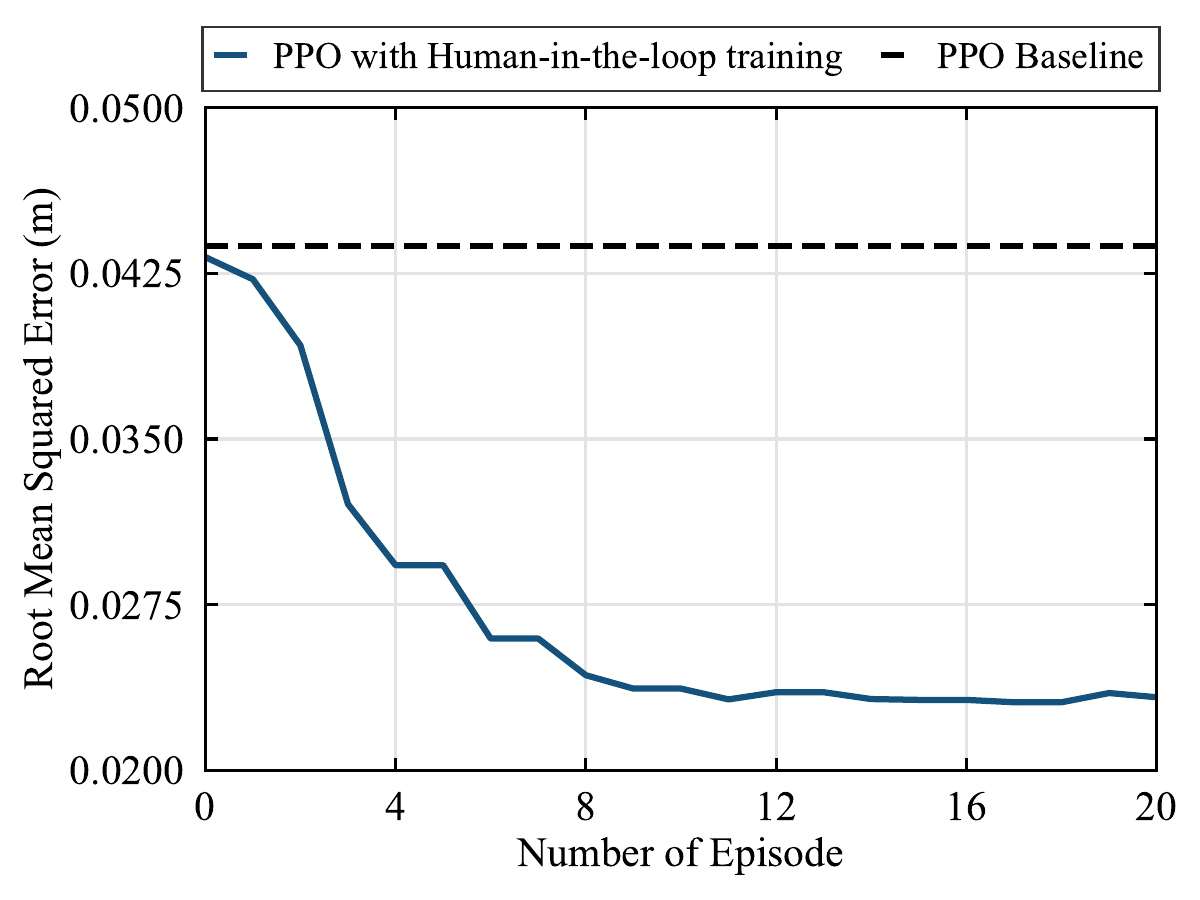}
\caption{Average \gls{rmse} in each training episode.}
\label{fig: reward_human}
\end{figure}

We first evaluate the performance of the \gls{arma} model. Specifically, the input length is set to 1000 ms and the length of predict horizon is set to 500 ms. The moving average order is set to $q = 2$ and the auto-regressive order is set to $p = 6$ with the lag sequence $[1,20,40,60,80,100]$. Table~\ref{teble_pre} presents the average \gls{rmse} for the position and orientation over the 500 ms period. We compare the proposed \gls{arma} model with two basedline prediction models, Informer and \gls{ar}~\cite{choi2012arma}. Our results show that the average \gls{rmse} is 0.00100 m for position and 0.00101 for orientation, with variances of  $5.264\times10^{-5} \text{m}$ for position and $5.650 \times 10^{-5}$ for orientation, demonstrating the superiority compared to other two algorithm. 


To evaluate the proposed DRL algorithm, we train the \gls{ppo} algorithm over 150 episodes for 10 times. The mean and the standard deviation of communication latency is set to $75$~ms and $12.5$~ms, respectively. The reward function weights are set to $w_1 = w_2 = w_3 = w_4 = -1$. 
As shown in Fig.~\ref{fig: reward}, the reward fluctuates for the first 30 episodes before reaching convergence with RMSE of 0.0443 m after 60 episodes of training. Then, we further do the human-in-the loop training based on the pre-trained convergences PPO baseline model for an additional 20 episodes. As shown in Fig.~\ref{fig: reward_human}, a further \gls{rmse} decreasing to 0.0241 m with 83.82\% gain is obtained compared with  PPO baseline model, which verified the effectiveness of our proposed trained algorithm.

\section{Conclusion}
In this work, we established a human-in-the-loop cross-system design framework to minimize the motion-to-photon latency on the modeling error of a real-world robotic arm and its matched virtual model in the Metaverse. We built a framework structure with separated display/control functions and human-in-the-loop \gls{drl} training process to improve the performance of the framework. The results our proposed method can significantly reduced 1) the \gls{mtp} latency between human motion and rendering feedback and 2) the \gls{rmse} between human motion and real-world remote devices.

\bibliographystyle{IEEEtran}
\bibliography{bib}

\vspace{12pt}
\end{document}